%% file: main.tex
\documentclass{article}

\usepackage[preprint]{neurips_2026}
\AtBeginDocument{\newgeometry{textheight=9.1in, textwidth=5.5in, top=1in,
  headheight=12pt, headsep=25pt, footskip=30pt}}

\usepackage[utf8]{inputenc}
\usepackage[T1]{fontenc}

\usepackage{hyperref}
\usepackage{url}
\usepackage{booktabs}
\usepackage{amsfonts}
\usepackage{amsmath}
\usepackage{amssymb}
\usepackage{nicefrac}
\usepackage{microtype}
\usepackage{graphicx}
\usepackage{multirow}
\usepackage{wrapfig}
\usepackage[table]{xcolor}
\usepackage{enumitem}
\usepackage{algorithm}
\usepackage{algpseudocode}

\input{macros}
\input{supp_labels}

\title{\methodname: Scene Composition with\\Instanced Neural Primitives}

\author{%
  William Koch \quad Amogh Joshi \quad Cyrus Vachha \quad
  Cheng Zheng \quad Felix Heide \\
  Princeton University \\
  \texttt{\{william.koch,amoghjoshi,cvachha,chengzh,fheide\}@princeton.edu}%
}

\begin{document}

\maketitle

\input{sections/00_abstract}
\input{sections/01_introduction}
\input{sections/02_related_work}

\input{sections/03_method}
\input{sections/04_experiments}
\input{sections/05_conclusion}

\bibliographystyle{plainnat}
\bibliography{main}

\end{document}

%% file: macros.tex
\newif\ifdraft
\draftfalse

\newcommand{\method}{\textsc{SCION}}
\newcommand{\methodname}{\method}
\newcommand{\ours}{Ours}
\newcommand{\ourssmall}{\ours-Small}
\newcommand{\ourslarge}{\ours-Large}

\newcommand{\splat}{splat}
\newcommand{\splats}{splats}

\newcommand{\cand}{primitive}
\newcommand{\cands}{primitives}
\newcommand{\Cand}{Primitive}

\newcommand{\inst}{instance}
\newcommand{\insts}{instances}

\newcommand{\nsplats}{S}

\newcommand{\ncands}{K}

\newcommand{\ninsts}{N}
\newcommand{\ntotal}{T}

\newcommand{\codebook}{\mathcal{C}}

\newcommand{\assign}{a}

\newcommand{\instpos}{\boldsymbol{t}}
\newcommand{\instrot}{\boldsymbol{R}}
\newcommand{\instscale}{\sigma}

\newcommand{\Loss}{\mathcal{L}}
\newcommand{\lrecon}{\Loss_{\mathrm{recon}}}

\newcommand{\lreg}{\Loss_{\mathrm{reg}}}
\newcommand{\lscale}{\Loss_{\mathrm{scale}}}
\newcommand{\lopacity}{\Loss_{\mathrm{opac}}}
\newcommand{\ladv}{\Loss_{\mathrm{adv}}}
\newcommand{\ltotal}{\Loss_{\mathrm{total}}}

\newcommand{\vs}{vs.\ }

\newcommand{\R}{\mathbb{R}}

\newcommand{\set}[1]{\mathcal{#1}}

\RequirePackage[table]{xcolor}

\definecolor{bestcell}{HTML}{D8F0D2}    %
\definecolor{secondcell}{HTML}{FCE9C2}  %
\newcommand{\best}[1]{\cellcolor{bestcell}#1}
\newcommand{\second}[1]{\cellcolor{secondcell}#1}

\newcommand{\figref}[1]{Fig.~\ref{fig:#1}}
\newcommand{\tabref}[1]{Tab.~\ref{tab:#1}}
\newcommand{\secref}[1]{Sec.~\ref{sec:#1}}
\newcommand{\eqnref}[1]{Eq.~\ref{eq:#1}}

%% file: sections/00_abstract.tex
\begin{abstract}
Real-world scenes are compositional: bricks, blades of grass, pebbles, and tree leaves recur across human-built and natural environments. Existing neural scene representations model these elements independently. Most 3D Gaussian Splatting and follow-up abstraction and compression methods treat each element as unique, fitting millions of independent Gaussians per scene. Prior methods like Splat and Replace fit template objects, but they require mostly manual selection of repeated elements. As a result, these representations store redundant parameters and provide weak manipulation handles for downstream tasks.
We introduce \method{}, a hierarchical compositional scene representation that replaces independent Gaussians with a compact vocabulary of reusable \cands{} and lightweight world-space \insts{} that place transformed copies throughout the scene. We fit this representation to multi-view captures via a joint optimization over discrete and continuous scene parameters, combining two-level densification over \splats{} and \insts{} with an adversarial loss that preserves detail across shared \cands{}. 
The recovered structure yields a compact, controllable representation while maintaining high quality even at 1.2 MB. \method{} achieves rate-distortion favorable to existing Gaussian compression methods, and it enables instance-level scene editing and animation without retraining. Our results show that neural scene representations need not memorize scenes as independent \cands{}; they can discover reusable parts. Project webpage: \url{https://light.princeton.edu/SCION}
\end{abstract}

%% file: sections/01_introduction.tex
\section{Introduction}
\label{sec:introduction}

The natural and built world is full of repetition. A meadow is a few
species of grass blade copied a million times; a brick wall is one
brick under different poses; a forest canopy is a handful of leaf
shapes scattered across branches. Traditionally, in computer graphics, we
therefore do not manually handcraft every single leaf individually, but just
a handful of templates which we then instantiate using a particle system or
array~\citep{Prusinkiewicz1990TheAB,stava2014inverse}. This does more
than save labor: the resulting scenes are much easier to animate, are faster
to edit, and require significantly less memory.

3D reconstruction deals with the inverse problem: going from images to a 3D
representation. Neural radiance fields and related neural graphics primitives
shifted reconstruction toward photorealistic view
synthesis~\citep{mildenhallNeRFRepresentingScenes2020,mullerInstantNeuralGraphics2022}.
However, because these scenes are represented as neural fields rather than
physical structures, they are hard to integrate into graphics workflows.
3D Gaussian Splatting (3DGS)~\citep{kerbl3DGaussianSplatting2023} in turn
uses an explicit representation, which makes rendering and integration into
other tools easier and more efficient. Yet 3DGS fits millions of independent
Gaussians per scene, each carrying its own position, covariance, opacity, and
color. The reconstructions are accurate but they are also large, and
\emph{structureless}: nothing in the representation tells us when the ten
thousand \splats{} covering a lawn are \insts{} of the same few blades.

Lack of structure limits the downstream usefulness of these scenes in two
ways. First, it makes edits hard to express: an intuitive operation such as
placing grass on a surface requires manually instantiating and tuning
thousands of individual \splats{}, instead of inserting a few reusable
\cands{}. Second, it makes animation hard: the user must deform an
unstructured collection of Gaussians instead of transforming coherent
objects. Semantic segmentation can isolate large objects, but it does not
recover the fine-grained repeated structure of leaves, grass, or other
repetitive elements.

Recovering structure requires a representation that discovers repetition.
Neural rendering has largely left this problem untouched. Prior
codebook-style methods target
compression~\citep{takikawaVariableBitrateNeural2022,leeCompact3DGaussian2024,durvasulaContraGSCodebookCondensedTrainable2025}:
they share parameters across distinct \splats{} rather than building shared
\cands{}. They identify repetition in parameters, not in geometry.

Explicit repetition has mostly appeared in analysis or exemplar pipelines.
Regularity detection analyzes input point clouds or meshes to discover
similarity-group repetitions, rather than reconstructing them from
images~\citep{paulyDiscoveringStructuralRegularity2008}. The closest 3DGS
work in this area, Splat and Replace, starts from an already trained 3DGS scene and
relies on human-guided instance selection~\citep{violante2025splat}. This makes it suitable for larger
annotated objects, but difficult to scale to dense fine-grained repetitions.

Identifying \textit{repetition} creates two optimization challenges: choosing
the unit of reuse, and tolerating variation across reused elements. A scene
element can be represented as one \inst{} of a large \cand{} containing many \splats{}, or as many
\insts{} of a smaller \cand{} with fewer \splats{}; the latter uses
parameters more efficiently and yields a more structured representation, but
is harder to optimize. This is a minimum description length problem. At the same time, real \insts{} of the same element
are not identical, so direct pixel-wise sharing drives templates toward
blurry averages and erases high-frequency detail. It also means that there is no optimal solution to our minimum description length problem.

\begin{figure}[t]
  \centering
  \includegraphics[width=\linewidth]{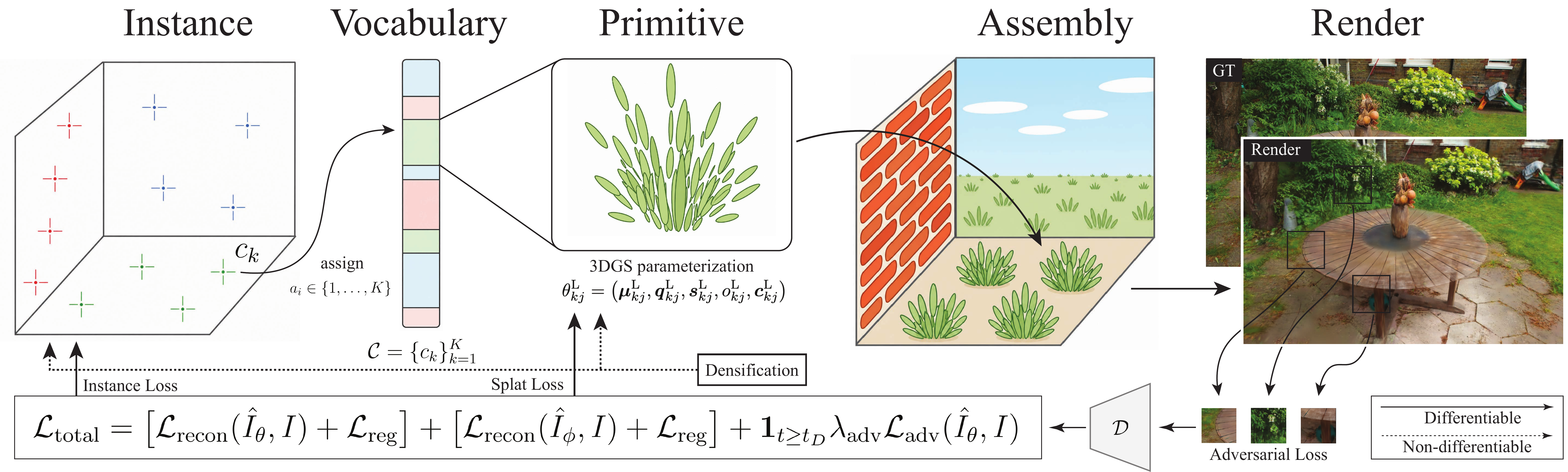}
  \vspace{-3mm}
  \caption{\textbf{Overview of \method{}.} We initialize \insts{}
  from a sparse point cloud, group them into reusable \cands{}, assemble
  the selected local \splats{} for rendering, and update the representation
  through image losses, projected adversarial supervision, and active
  densification.}
  \label{fig:overview}
\end{figure}

We introduce \method{}, which recovers a scene's \cand{} decomposition
end-to-end from multi-view captures (\figref{overview}). \method{}
represents a scene as a small library
$\codebook = \{c_k\}_{k=1}^{\ncands}$ of reusable \cands{}, each being a bundle
of 3D Gaussians in a local frame, and a set of \insts{}: lightweight
world-space placements with pose $(\instpos, \instrot)$, scale
$\instscale$, opacity $\alpha$, and assignment
$\assign_i \in \{1, \dots, \ncands\}$. The representation scales primarily 
with the number of sparse lightweight instances $\ninsts$ rather than the 
number of \cands{} $\ncands$ or \splats{} $\nsplats$. This leads to better rate--distortion tradeoffs.
The factorization into \cands{} furthermore gives the scene direct editing and
animation handles. 
Recovering this factorization is challenging: each \cand{} aggregates
gradients from all referencing \insts{}, and averaged reconstruction losses
(e.g., L1) drive templates toward blurry means, erasing high-frequency
detail. We address this with \cand{}-local \splat{} birth and relocation,
world-space \inst{} pruning, cloning, and subdivision, together with
an adversarial patch loss that preserves per-instance sharpness while still
enabling shared structure.

We make the following contributions in this work
\begin{itemize}
  \item We introduce \method{}, a hierarchical representation learning
        method for compositional scene reconstruction that replaces
        independent Gaussians with a reusable vocabulary of
        \cands{} and lightweight transformed \insts{}. The resulting 3D
        scenes are compact, structured, and directly manipulable.

  \item We learn \textit{repetitions}
        with new densification strategies for \cand{}-local \splats{} and
        world-space \insts{}, and integrate them in a joint optimization
        procedure that preserves sharpness across \inst{} aggregation.

  \item We demonstrate that the recovered structure has favorable
        rate--distortion tradeoffs over Gaussian compression methods at
        matched storage, while additionally enabling \inst{}-level editing
        and animation without retraining.
\end{itemize}

%% file: sections/02_related_work.tex
\section{Related work}
\label{sec:related}

\method{} lies at the intersection of neural rendering,
repeated-structure discovery, and controllable scene representations. We
review each of these areas: Prior compression methods largely preserve independent
\splats{} while encoding them more efficiently; pattern-recognition methods
often recover repeated objects under stronger priors; and editing methods
usually add controls on top of unstructured \cands{}. In contrast,
\method{} makes repetition part of the representation itself by learning
shared geometric \cands{} and lightweight \inst{} assignments directly from
multi-view images.

\paragraph{Neural rendering \& compression.}
3D Gaussian Splatting~\citep{kerbl3DGaussianSplatting2023} represents scenes with millions of independently parameterized anisotropic elements. Many variants improve this representation by modifying underlying geometry, optimization, or organization, e.g., surface-aligned disks or surfels~\citep{huang20242d,Dai2024GaussianSurfels}, mesh-attached Gaussians~\citep{guedon2023sugar}, hierarchical levels of detail~\citep{hierarchicalgaussians24}, or sampling-based densification and pruning~\citep{kheradmand3DGaussianSplatting2024,kimMetropolisHastingsSampling3D2025,durvasulaContraGSCodebookCondensedTrainable2025}. However, these methods largely preserve the per-\splat{} parameterization.

A large body of work compresses 3DGS by improving parameter encoding, entropy modeling, quantization, or predictive coding while keeping the underlying \splat{} representation fixed~\citep{luScaffoldgsStructured3d2024,chenHACHashgridAssisted2024,zhanCAT3DGSContextadaptiveTriplane2025,wangContextGSCompact3D2024,chenHAC100XCompression2025,liu3DGaussianSplatting2025,liuCompGSEfficient3D2024,xuLearningHierarchicalSparse2026,pcgs2025,liuFeedForward3DGaussian2025}. Other approaches modify the representation more directly through codebooks, residual quantization, distilled appearance, 2D projections, or alternative primitive elements such as textured surfels and billboards~\citep{Niedermayr_2024_CVPR,leeCompact3DGaussian2024,leeOptimizedMinimal3D2025,fan2024lightgaussian,morgenstern2024compact,rongNexelsNeurallyTexturedSurfels2025,svitovBillBoardSplattingBBSplat2025}.

Related compression ideas also appear in NeRFs, neural fields, point clouds, and 3D-aware generative models, where shared networks, grids, latent codebooks, or local decoders reduce storage~\citep{mildenhallNeRFRepresentingScenes2020,mullerInstantNeuralGraphics2022,tangNeuralgsBridgingNeural2026,watanabeNeuralGaborSplatting2026,maPixelCodecAvatars2021,vandenoordNeuralDiscreteRepresentation2017,takikawaVariableBitrateNeural2022,youPatchBasedDeepAutoencoder2021,liCompressingVolumetricRadiance2023,schwarz2020graf,Niemeyer2020GIRAFFE}. Across these methods, the atomic unit is still typically an independently stored \cand{}, feature, or latent code. In contrast, \method{} reduces the number of independent parameters by sharing concrete \emph{geometric} \cands{} across many \insts{} through a discrete vocabulary $\codebook$, improving compression while preserving perceived quality across scenes.

\paragraph{Instance and pattern recognition.}
Recovering repeated structure has a long history, but most previous methods assume existing geometry, manual inputs, or strong structural priors. Classical regularity detection fits similarity-group lattices and representative elements to point clouds or meshes~\citep{paulyDiscoveringStructuralRegularity2008}, while inverse procedural modeling fits authored grammars to curated exemplars such as polygonal tree scans~\citep{stava2014inverse} or pre-segmented part hierarchies~\citep{liGRASSGenerativeRecursive2017}. Splat and Replace is closer to ours in using 3DGS, but starts from a trained scene and human-guided instance selection, then registers and fuses object-scale repeats post hoc~\citep{violante2025splat}. Gaussian splat instantiation has also been explored for tiling, but not reconstruction~\citep{zengGSWTGaussianSplatting2025}.

Recent multi-object reconstruction aggregates unaligned views of repeated objects, but still requires per-object segmentation~\citep{wu2026jrm}. Slot-based methods discover compositional radiance fields without masks~\citep{yuUnsupervisedDiscoveryObject2022}. Inverse-rendering approaches decompose scenes into geometry, materials, and lighting at urban scale~\citep{wangNeuralFieldsMeet2023}. However, both operate at object-level granularity. Feature uplifting transfers 2D foundation features into 3D for grouping~\citep{marrieLUDVIGLearningfreeUplifting2025,simeoniDINOv32025}, inheriting category bias and object-scale receptive fields. Category-specific pipelines further restrict scope to a single class, such as botanical trees~\citep{liLearningReconstructBotanical2021}.

In contrast, \method{} learns a \cand{} vocabulary $\codebook$ and discrete \inst{} assignments end-to-end from multi-view images. It requires no grammar, exemplar, or category prior, and as such, generalizes to any scene with repetitive structures in the wild.

\paragraph{Controlling neural representations.}
Recent work adds control to 3DGS scenes by augmenting independently parameterized \splats{}. Editing methods support brush- or patch-based authoring with diffusion inpainting~\citep{pandey2025painting}, test-time latent editors~\citep{zhengSplatPainterInteractiveAuthoring2025}, or planar-prior anchoring for cleaner edits in textureless regions~\citep{tibenskyAnchorGSAnchoringGaussian2025}.

For motion, graph-based hierarchies infer parent--child relations for ARAP-style deformation~\citep{zheng2025heir}, while dynamic 3DGS methods deform canonical Gaussians with MLP or HexPlane-style fields~\citep{yang2024deformable,wu20244d}, but are typically not directly editable. Other controllable methods drive Gaussians using sparse control graphs, rigged face meshes, skinning, tracking, modal bases, or motion priors distilled from diffusion and spectral models~\citep{huang2024sc,qian2024gaussianavatars,wimmer2025gaussianstolife,li2024generative}. These methods are often object-centric and add control on top of independent splats, rather than producing shared structures that are naturally animatable.

\method{} does not introduce a dedicated editing or animation system. Instead, its instance-based representation enables both implicitly. Edits or motions applied to a \cand{} propagate to all associated \insts{} without per-splat handles. This is particularly effective in highly repetitive regimes, such as foliage, masonry, or tiled surfaces, where object-centric control is ill-defined.

%% file: sections/03_method.tex
\section{Learning the scene composition}
\label{sec:method}

\method{} learns a factorized Gaussian scene representation from posed
training images. The scene is stored as a reusable vocabulary of local
\cands{} and a larger set of world-space \insts{}: each \inst{} selects
one \cand{} and places a transformed copy into the scene. We jointly optimize the shared \cand{} parameters and
the \inst{} placements with both image and adversarial
supervision, as well as a new vocabulary-aware density optimization.

\begin{figure}[t]
  \centering
  \includegraphics[width=\linewidth]{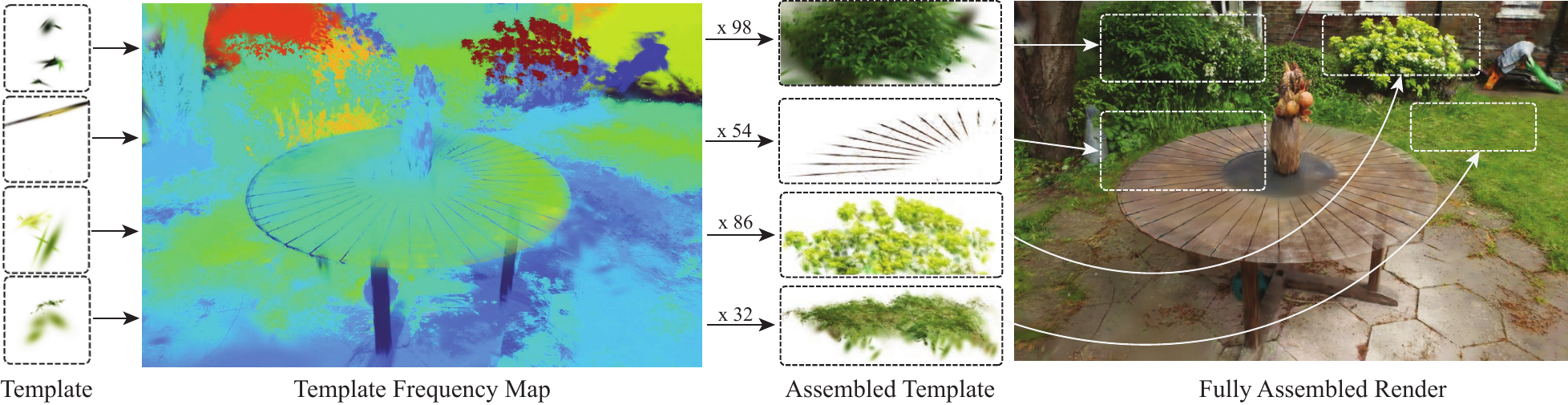}\\[-0.5mm]
  {\footnotesize\setlength{\unitlength}{\linewidth}\begin{picture}(1,0.025)
    \put(0.030,0.01){\makebox(0,0){(a)}}\put(0.269,0.01){\makebox(0,0){(b)}}
    \put(0.568,0.01){\makebox(0,0){(c)}}\put(0.844,0.01){\makebox(0,0){(d)}}
  \end{picture}}
  \vspace{-4mm}
  \caption{\textbf{Neural \cand{} assembly.} The (a) learned \cands{} are (b) instantiated multiple times throughout the scene, resulting in (c) assembled \cand{} structures that (d) construct the scene.}
  \label{fig:process}
  \vspace{-2mm}
\end{figure}

\subsection{Problem setup and representation}
\label{sec:method-hierarchy}

Given posed RGB images $\{I_v\}$ with calibrated cameras $\{\pi_v\}$, our
goal is to recover a compact scene representation that can render novel
views while exposing repeated structure. \method{} represents the scene by
a vocabulary $\codebook=\{c_k\}_{k=1}^{\ncands}$ and a set of \insts{}
$\Phi=\{\phi_i^{\mathrm{W}}\}_{i=1}^{\ninsts}$. \Cand{} $c_k$ is a
reusable local-frame template containing $S_k$ \splats{}. The $j$th
\splat{} in $c_k$ has parameters $\theta_{kj}^{\mathrm{L}}$, and the
$i$th \inst{} has parameters $\phi_i^{\mathrm{W}}$ plus a discrete
assignment $\assign_i$ selecting its template:
\[
\begin{aligned}
  \theta_{kj}^{\mathrm{L}}
    &=
    \bigl(
      \boldsymbol{\mu}_{kj}^{\mathrm{L}},
      \boldsymbol{q}_{kj}^{\mathrm{L}},
      \boldsymbol{s}_{kj}^{\mathrm{L}},
      o_{kj}^{\mathrm{L}},
      \boldsymbol{c}_{kj}^{\mathrm{L}}
    \bigr), \\
  c_k
    &=
    \{\theta_{kj}^{\mathrm{L}}\}_{j=1}^{S_k}, \\
  \phi_i^{\mathrm{W}}
    &=
    \bigl(
      \boldsymbol{\mu}_{i}^{\mathrm{W}},
      \boldsymbol{q}_{i}^{\mathrm{W}},
      s_i^{\mathrm{W}},
      o_i^{\mathrm{W}},
      \assign_i
    \bigr).
\end{aligned}
\label{eq:codebook-params}
\]
The superscripts mark the frame: local \cand{} coordinates ($\mathrm{L}$)
or world coordinates ($\mathrm{W}$). Each local \splat{} carries a
position $\boldsymbol{\mu}_{kj}^{\mathrm{L}}\in\R^3$, a unit-quaternion
rotation $\boldsymbol{q}_{kj}^{\mathrm{L}}\in S^3$, an anisotropic
log-scale $\boldsymbol{s}_{kj}^{\mathrm{L}}\in\R^3$, an opacity logit
$o_{kj}^{\mathrm{L}}\in\R$, and an RGB color
$\boldsymbol{c}_{kj}^{\mathrm{L}}\in\R^3$. Each \inst{} carries a
world-frame position $\boldsymbol{\mu}_i^{\mathrm{W}}\in\R^3$ and
rotation $\boldsymbol{q}_i^{\mathrm{W}}\in S^3$, a scalar log-scale and
opacity logit $s_i^{\mathrm{W}}, o_i^{\mathrm{W}}\in\R$, and a discrete
\cand{} index $\assign_i\in\{1,\ldots,\ncands\}$.
We disable view-dependent appearance: We store
only $\boldsymbol{c}_{kj}^{\mathrm{L}}$, the RGB logit for the
view-independent zeroth-order spherical-harmonic coefficient. Higher-order
spherical-harmonic coefficients are omitted. When a later equation is
unambiguous, we omit the frame superscript.

Let $n_k=|\{i:\assign_i=k\}|$ be the number of \insts{} assigned to
\cand{} $k$.
Rendering composes the local \splats{} of each assigned \cand{} with the
\inst{} transform. For each \inst{} $i$ and each local \splat{}
$j \in \{1,\ldots,S_{\assign_i}\}$ in its assigned \cand{}, we assemble a
world-frame Gaussian
\begin{align}
  \tilde{\boldsymbol{\mu}}_{ij}^{\mathrm{W}}
    &=
    \boldsymbol{\mu}_i^{\mathrm{W}}
    +
    \boldsymbol{q}_i^{\mathrm{W}}
    \!\left(e^{s_i^{\mathrm{W}}}
    \boldsymbol{\mu}_{\assign_i j}^{\mathrm{L}}\right), \\
  \tilde{\boldsymbol{s}}_{ij}^{\mathrm{W}}
    &=
    \boldsymbol{s}_{\assign_i j}^{\mathrm{L}}
    + s_i^{\mathrm{W}}\mathbf{1}, \\
  \tilde{\boldsymbol{q}}_{ij}^{\mathrm{W}}
    &=
    \boldsymbol{q}_i^{\mathrm{W}}
    \otimes
    \boldsymbol{q}_{\assign_i j}^{\mathrm{L}}, \\
  \tilde{o}_{ij}^{\mathrm{W}}
    &=
    o_{\assign_i j}^{\mathrm{L}} + o_i^{\mathrm{W}},
    \qquad
  \tilde{\boldsymbol{c}}_{ij}^{\mathrm{W}}
    =
    \boldsymbol{c}_{\assign_i j}^{\mathrm{L}} .
  \label{eq:assembly}
\end{align}
where $\boldsymbol{q}(\cdot)$ denotes the rotation of a vector by a quaternion, and $\otimes$ is quaternion composition. 
A render materializes
$\ntotal=\sum_{i=1}^{\ninsts}S_{\assign_i}=\sum_k n_kS_k$ Gaussians as
transient tensors.

This factorization gives \method{} a smaller representation space than a
flat 3DGS scene~\citep{kerbl3DGaussianSplatting2023}, which stores a
separate parameter vector for every rendered Gaussian. If
$\bar S=\frac{1}{\ncands}\sum_{k=1}^{\ncands}S_k$ is the average number
of local \splats{} per \cand{}, then the representation space complexity $\mathcal{O}(\ncands \bar S + \ninsts)$.
The first term stores the vocabulary, and the second stores the
world-frame \inst{} parameters and assignment indices. By contrast, an
unshared representation with the same assembled Gaussians has
$M_{\mathrm{flat}}=\mathcal{O}(\sum_i S_{\assign_i})$.

The exact count only differs by fixed parameter dimensions. Let
$P_{\splat}=14$ be the number of parameters per local \splat{}, and let $P_{\inst}=10$ be the
number of parameters per \inst{}, not counting the integer
assignment. Note that while the rasterizer still grows with the number of assembled Gaussians, the stored representation only has a total number of parameters
$B_{\mathrm{ours}} = P_{\splat}\sum_{k=1}^{\ncands} S_k + P_{\inst}\ninsts$.
The compression ratio relative to an unshared representation with the
same assembled Gaussians is
$\mathrm{CR} = B_{\mathrm{ours}} / \bigl(P_{\splat}\sum_k n_kS_k\bigr)$.

We initialize \insts{} from the sparse COLMAP point
cloud~\citep{schoenberger2016sfm} and use frozen DINOv3
features~\citep{simeoniDINOv32025} on the training images to assign them to
\cands{}. During scene reconstruction, we keep this assignment fixed. Initialization details are given in \secref{appendix-initialization}.

\subsection{Learning objective}
\label{sec:method-training}

To ensure stability, we optimize \splats{} and \insts{} in two separate phases of an iteration. To avoid coupled gradients between \cand{} and \inst{} parameters, each training iteration renders the same view twice. Let $\theta$ be the
\cand{}'s \splat{} parameters and let
$\phi$ be the \inst{} parameters. During the \splat{} pass we detach $\phi$, and vice-versa in the \inst{} pass:
\[
  \hat I_{\theta}=R(\theta,\operatorname{sg}(\phi)),
  \qquad
  \hat I_{\phi}=R(\operatorname{sg}(\theta),\phi).
  \label{eq:scoped-gradients}
\]
Here $R$ is the Gaussian rasterizer and $\operatorname{sg}$ stops
gradients. The two RGB images match; only the autograd graph differs.

For the image loss, we follow the 3DGS photometric
objective~\citep{kerbl3DGaussianSplatting2023}, which combines L1 with
SSIM~\citep{wang2004image}, but add a scheduled
LPIPS~\citep{zhang2018unreasonable} term:
\[
  \lrecon^{(t)}(\hat I, I)
  =
  (1-\lambda_{\mathrm{ssim}})\|\hat I-I\|_1
  + \lambda_{\mathrm{ssim}}\,(1-\mathrm{SSIM}(\hat I,I))
  + \mathbf{1}_{\mathrm{lpips}}(t)\,
    \lambda_{\mathrm{lpips}}\,
    \mathrm{LPIPS}_{\mathrm{vgg}}(\hat I,I).
  \label{eq:recon-loss}
\]

We add a regularization term that combines two standard 3DGS-style
\splat{} regularizers with three terms specific to the
vocabulary--\inst{} factorization:
\begin{equation}
  \begin{aligned}
  \lreg
  &=
  \lscale + \lopacity
  + \lambda_{\mathrm{anch}} \frac{1}{\ninsts}\sum_i |s_i^{\mathrm{W}}|
  + \lambda_{\mathrm{cap}} \sum_i[s_i^{\mathrm{W}}-\tau_{\mathrm{cap}}]_+^2 \\
  &\quad
  + \lambda_c
  \frac{1}{|\codebook_{\geq 2}|}\sum_{k\in\codebook_{\geq 2}}
  \frac{1}{S_k^+}\sum_{j\in\set{A}_k}
  \|\boldsymbol{\mu}_{kj}^{\mathrm{L}}
    -\bar{\boldsymbol{\mu}}_k^{\mathrm{L}}\|_2^2 .
  \end{aligned}
  \label{eq:reg-loss}
\end{equation}
$\lscale$ penalizes anisotropic local-frame \splat{} scales and
$\lopacity$ caps large local opacity logits; both follow standard 3DGS
practice and are written out in \secref{appendix-reg}. $\set{A}_k$ is
the active subset of \cand{} $k$, $S_k^+=|\set{A}_k|$, and $\codebook_{\geq 2}$ is
the set of \cands{} with at least two active \splats{}. The remaining
three terms reflect the vocabulary--\inst{} factorization. The
$\lambda_{\mathrm{anch}}$ term anchors each \inst{}'s log-scale near its
initialization, since \insts{} should place templates at their seeded
footprint rather than rescale them. The $\lambda_{\mathrm{cap}}$ barrier caps
positive \inst{} log-scales to prevent runaway growth. The $\lambda_c$
term keeps each \cand{}'s active \splats{} close to their local-frame
mean, preventing template drift away from the \cand{}'s local
neighborhood.

The full learning objective is
\[
  \ltotal^{(t)}
  =
  \bigl[\lrecon^{(t)}(\hat I_{\theta},I)+\lreg\bigr]
  +
  \bigl[\lrecon^{(t)}(\hat I_{\phi},I)+\lreg\bigr]
  + \mathbf{1}_{t\ge t_D}\lambda_{\mathrm{adv}}\ladv(\hat I_{\theta}, I).
  \label{eq:total-loss}
\]
Photometric losses can average detail across repeated \insts{}, so after a
warmup step $t_D$ we add an adversarial loss~\citep{goodfellow2014generative}
on the \splat{}-pass render. Similar to patch-level adversarial
supervision for 3D-aware rendering~\citep{schwarz2020graf}, the
discriminator compares randomly cropped real and rendered patches.
We only apply this to the \splat{}-pass render, so the discriminator learns to preserve sharpness across \inst{} aggregation.

\subsection{Vocabulary-aware density learning}
\label{sec:method-strategies}

The default method keeps the \cand{} set fixed after initialization.
Our densification modifies only local \splats{} inside \cands{} and
world-space \insts{}. Both
operators run on a refinement schedule and read gradients accumulated over
the previous window.

\paragraph{Gradient accumulation.}
The rasterizer provides screen-space absolute gradients at the projected
means of assembled world Gaussians. Over a refinement window $\mathcal W$,
one \cand{}'s \splat{} appears through many \insts{}, so its signal is a
scatter-add over all rendered copies:
\begin{equation}
  \bar g_r
  =
  \frac{
    \sum_{v \in \mathcal W}
    \sum_{(i,j):\,r=(\assign_i,j)}
    \|\nabla_{\pi_v(\tilde{\boldsymbol{\mu}}_{ij}^{\mathrm{W}})}
      \ltotal^{(v)}\|_2}
  {
    \sum_{v \in \mathcal W}
    \sum_{(i,j):\,r=(\assign_i,j)} 1
  }.
  \label{eq:splat-grad-accum}
\end{equation}
Instance gradients use the same scatter-add pattern, but group by \inst{}:
\begin{equation}
  \bar h_i
  =
  \frac{
    \sum_{v \in \mathcal W}
    \sum_{j=1}^{S_{\assign_i}}
    \|\nabla_{\pi_v(\tilde{\boldsymbol{\mu}}_{ij}^{\mathrm{W}})}
      \ltotal^{(v)}\|_2}
  {
    \sum_{v \in \mathcal W} S_{\assign_i}
  }.
  \label{eq:inst-grad-accum}
\end{equation}

\paragraph{Splat density optimization.}
We use the MH-3DGS birth rule~\citep{kimMetropolisHastingsSampling3D2025}
as a row-allocation mechanism for the vocabulary, not as a global Gaussian
sampler. We scope every proposal to one \cand{}: we fill a dead row in
$c_k$ with an opacity-weighted alive donor from $c_k$, and write any new
child proposed from a parent in $c_k$ back into $c_k$. This preserves
the DINO-initialized \cand{} identity and avoids cross-template drift.

Beyond this scoping, we make three vocabulary-specific adaptations. (i)
The parent importance score is z-scored within each \cand{}
(\eqnref{mh-importance}), so births respond to where a template is
under-fit relative to its own rows rather than to the total rasterizer
signal aggregated across many instances. (ii) Acceptance is divided by a
same-\cand{} voxel density (\eqnref{mh-accept}) to prevent births from
piling up in one part of a template. (iii) When a row is copied into $\nu$
overlapping instances, opacity and scale are corrected to preserve
alpha-compositing mass (\eqnref{mass-relocation}). Jitter,
visit-count damping, and coarse/fine proposal annealing follow MH-3DGS;
the full row update, free-list handling, and accumulator reset are given
in Algorithm~\ref{alg:splat-mh}, and the three adaptation equations are
collected in \secref{appendix-mh-splat}.

\paragraph{Instance density optimization.}
Instance densification changes world coverage while keeping the same
\cand{} set. At each firing, we first remove instances whose opacity is
below threshold or whose parameters are non-finite. We rank the remaining
instances by $\bar h_i$ from \eqnref{inst-grad-accum}.

High-gradient instances split into two cases. We clone a small instance
when its linear scale $e^{s_i^{\mathrm{W}}}$ falls below the split
threshold:
\begin{equation}
  \boldsymbol{\mu}_{i'}^{\mathrm{W}}
  =
  \boldsymbol{\mu}_i^{\mathrm{W}}
  + \eta_{\mathrm{clone}}e^{s_i^{\mathrm{W}}}\xi,
  \qquad
  \bigl(
    \boldsymbol{q}_{i'}^{\mathrm{W}},
    s_{i'}^{\mathrm{W}},
    o_{i'}^{\mathrm{W}},
    \assign_{i'}
  \bigr)
  =
  \bigl(
    \boldsymbol{q}_{i}^{\mathrm{W}},
    s_i^{\mathrm{W}},
    o_i^{\mathrm{W}},
    \assign_i
  \bigr),
  \quad
  \xi\sim\mathcal N(0,I).
  \label{eq:inst-clone}
\end{equation}
We subdivide a large instance into $M$ children. Children inherit the
parent rotation, assignment, and opacity; only position and scale change:
\begin{equation}
  \boldsymbol{\mu}_{i,m}^{\mathrm{W},\prime} =
  \boldsymbol{\mu}_i^{\mathrm{W}}
  + \eta_{\mathrm{sub}}e^{s_i^{\mathrm{W}}}\epsilon_m,
  \qquad
  s_{i,m}^{\mathrm{W},\prime} =
  s_i^{\mathrm{W}} - \frac{1}{3}\log M,
  \qquad
  o_{i,m}^{\mathrm{W},\prime} = o_i^{\mathrm{W}},
  \quad
  \epsilon_m\sim\mathcal N(0,I).
  \label{eq:inst-densify}
\end{equation}
This differs from \splat{} splitting. Instance children occupy different
world positions, so they should not use the alpha-compositing correction
from \eqnref{mass-relocation}. Reducing opacity here, in addition to our
rescaling of instances, would remove rendered mass from the scene. We cap both clone count and subdivide-parent count per
firing, selecting the highest-gradient eligible instances first, and reset
the accumulators after each update. The detailed procedure is in
Algorithm~\ref{alg:inst-default}.

%% file: sections/04_experiments.tex
\section{Experiments}
\label{sec:experiments}

We evaluate \method{} as a compact and meaningful scene representation. We trade per-pixel fidelity for reusable structures with direct manipulation handles. We showcase the validity of our method in two ways: quantitative evaluation of reconstruction quality under matched storage budgets, as well as qualitative evaluation, showcasing the meaningfulness with downstream applications.

\subsection{Evaluation protocol}
\label{sec:experiments-setup}

We evaluate on the Mip-NeRF 360~\citep{barron2022mip},
Tanks \& Temples~\citep{Knapitsch2017}, Deep
Blending~\citep{DeepBlending2018}, and
Synthetic-NeRF~\citep{mildenhallNeRFRepresentingScenes2020} benchmarks
using the standard training/test camera splits. Unless otherwise noted,
\method{} is trained for 30k iterations. We report PSNR,
SSIM~\citep{wang2004image},
LPIPS~\citep{zhang2018unreasonable}, and on-disk model
size in MB. Full per-benchmark results are deferred to
\tabref{main_results} in the supplementary material.

\subsection{Matched storage at 1-2 MB}
\label{sec:experiments-matched-storage}

\begin{wraptable}[12]{r}{0.5\linewidth}
  \vspace{-\intextsep}
  \centering
  \caption{\textbf{Matched-storage comparison on MipNeRF360.} Each
  baseline is rerun at the lowest available compression point point near our method.}
  \label{tab:matched_budget}
  \vspace{2mm}
  \small
  \setlength{\tabcolsep}{4pt}
  \input{tables/table_6_matched_budget}
\end{wraptable}
We compare how different representations behave around a 1.5\,MB target on
MipNeRF360, and qualitatively compare whether they preserve repeated structure at
that scale. We rerun state-of-the-art 3DGS compression baselines at the lowest available
operating points near our target: ContextGS~\citep{wangContextGSCompact3D2024},
HAC++~\citep{chenHAC100XCompression2025}, and
OMG~\citep{leeOptimizedMinimal3D2025}. We do not include methods that don't provide budget levers, or cannot be reasonably compressed below 2\,MB such as PCGS~\citep{pcgs2025}. \tabref{matched_budget} reports the model size and quality metrics, averaged over MipNeRF360~\citep{barron2022mip}.

\begin{figure}[t]
  \centering
  \IfFileExists{figures/qualitative_compression_eval.pdf}{%
    \includegraphics[width=\linewidth]{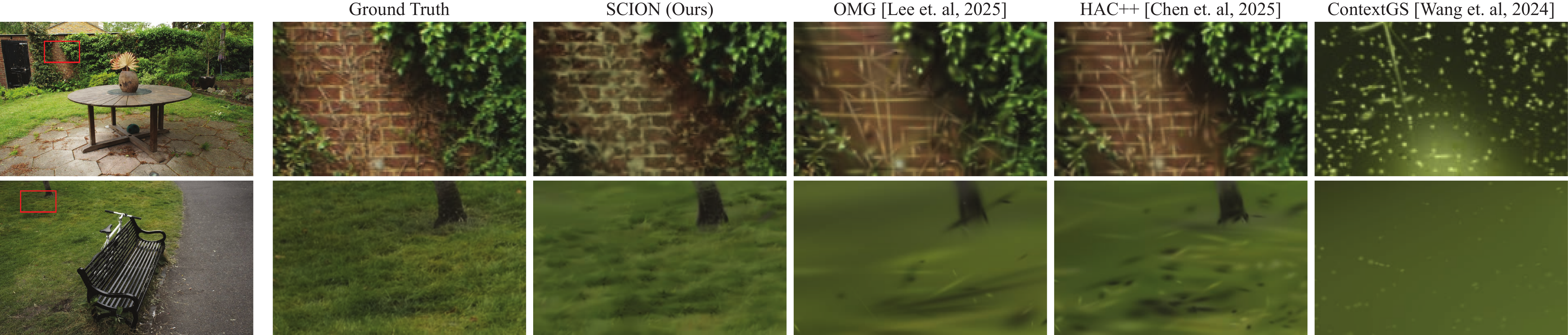}%
  }{%
    \fbox{\parbox{0.9\linewidth}{\centering\vspace{2em}%
    \vspace{2em}}}%
  }
  \vspace{-3mm}
  \caption{\textbf{Qualitative matched-storage comparison.} At
  approximately 1.5\,MB, \method{} retains coherent repeated structure as each reused \cand{} is represented once and instantiated many
  times.}
  \label{fig:compression_eval}
\end{figure}

\figref{compression_eval} shows how each representation spends the same
budget. Parameter-compression methods preserve a conventional 3DGS
parameterization and encode independent Gaussians. \method{} instead
stores a reusable \cand{} vocabulary plus many cheap \insts{}, so
repeated scene elements remain coherent even when raw pixel metrics
understate the perceptual quality.

We also report the full-rate results of these methods, and others, in \tabref{main_results} which calibrates the cost of the factorization. \method{} is not a drop-in replacement for high-fidelity Gaussian reconstructions: unstructured 3DGS compression methods achieve substantially better image metrics, as they do not share geometric \cands{} across the scene.
For context, the best PSNR reported in \tabref{main_results} reaches
27.62\,dB / 0.237 LPIPS on MipNeRF360, 24.22\,dB / 0.190 LPIPS on
Tanks \& Temples, 30.21\,dB / 0.254 LPIPS on Deep Blending, and
33.80\,dB / 0.031 LPIPS on Synthetic-NeRF at 12.68, 5.18, 66.00,
and 68.46\,MB respectively. \ourssmall{} instead uses 1.20, 1.55,
0.73, and 0.16\,MB on the same benchmark groups, however, with 22.00\,dB /
0.432 LPIPS, 18.61\,dB / 0.416 LPIPS, 24.02\,dB / 0.442 LPIPS, and
24.39\,dB / 0.105 LPIPS. \method{} trades raw PSNR for a structured
representation that supports edits and animation no flat \splat{} model can
express (\secref{experiments-edit-animation}), while cutting storage by
two to three orders of magnitude. Our learning objective harms PSNR and SSIM metrics while producing significantly better qualitative results, as described in \secref{experiments-ablations}.

\subsection{Ablations}
\label{sec:experiments-ablations}

\begin{figure}[t]
  \centering
  \IfFileExists{figures/04_ablation_figure.pdf}{%
    \includegraphics[width=\linewidth]{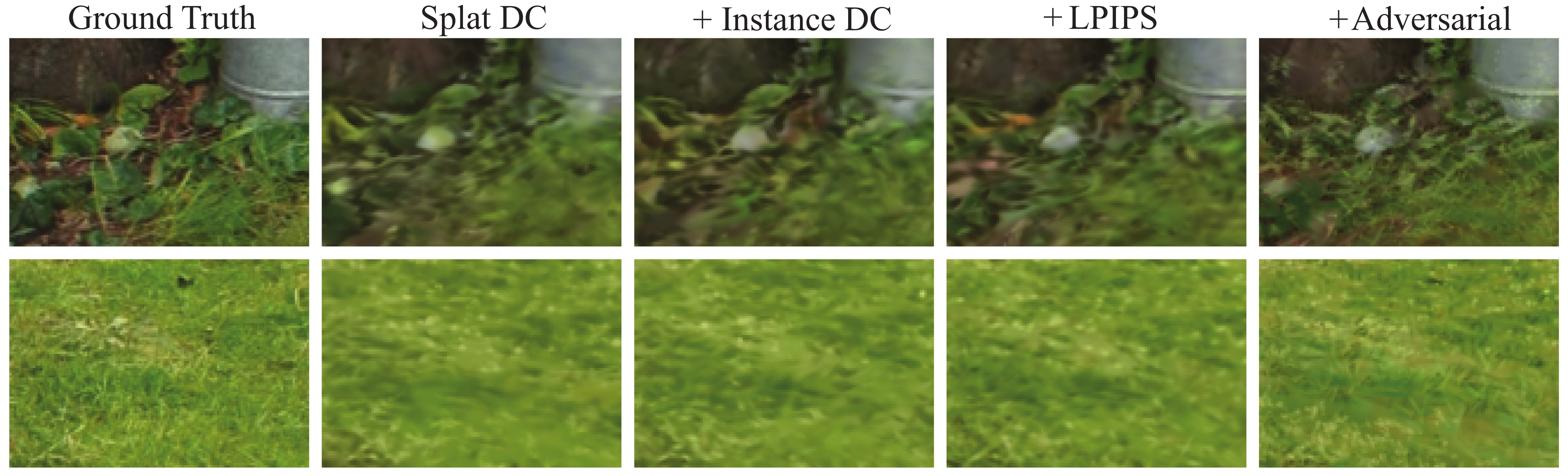}%
  }{%
    \fbox{\parbox{0.9\linewidth}{\centering\vspace{2em}%
    \vspace{2em}}}%
  }
  \vspace{-3mm}
  \caption{\textbf{Component ablation renders.} Example render of the
  \ourslarge{} variant of our method on the Garden scene from MipNeRF360.
  ``DC'' stands for density control.}
  \label{fig:component_ablation_renders}
\end{figure}

We show the impact of the different components of our learning objective both quantitatively in \tabref{component_ablation} and qualitatively in \figref{component_ablation_renders}. For the starting point, we disable LPIPS and adversarial loss, as well as disable densification control for instances. While it still uses a shared representation, densification only occurs on splat-level. We add the additional components of our methods cumulatively.

We observe a perceptual \vs reconstruction trade-off rather than a strict gain: the components improve LPIPS and qualitative results at a small cost in PSNR and SSIM. The biggest qualitative improvement is introduced by the adversarial training (see \figref{component_ablation_renders}). It is worth pointing out that the performance delta on LPIPS is larger by adding adversarial training than by adding an LPIPS loss - the very metric we are comparing against. This suggests that reference-based loss terms alone lead to a local optimization minimum. 

We share additional results on the impact of the number of \cands{} and initial \splats{} per \cand{} in \tabref{capacity_ablation}.

\begin{table}[t]
  \centering
  \caption{\textbf{Component ablation.} Rows cumulatively add the active
  \method{} components on the MipNeRF360 \texttt{garden} anchor scene. ``DC''
  stands for density control.}
  \label{tab:component_ablation}
  \small
  \setlength{\tabcolsep}{4pt}
  \input{tables/table_2_component_ablation}
\end{table}

\subsection{Editability and animation}
\label{sec:experiments-edit-animation}

\textbf{Editability.} The vocabulary representation exposes edit handles that
are absent from a flat set of independent \splats{}. We implement an
editor that lets users add, remove, or replace recovered \cands{} through
their \insts{}. Because the initial vocabulary assignment carries
instance-level DINO features, users can retrieve \cands{} with
text queries and insert additional instances into the scene. Additional
details and examples appear in \secref{appendix-edit-animation} and the
supplementary video.

\begin{figure}[t]
  \centering
  \IfFileExists{figures/06_animation.pdf}{%
    \includegraphics[width=\linewidth]{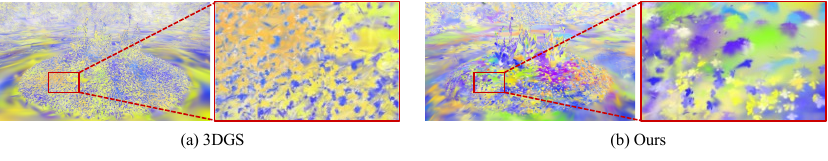}%
  }{%
    \fbox{\parbox{0.9\linewidth}{\centering\vspace{1.5em}%
    }}%
  }
  \vspace{-7mm}
  \caption{\textbf{Animation.} Flow maps for the same wind animation applied to
  (a) 3DGS and (b) \method{}. Per-splat transformations in 3DGS produce
  incoherent local motion, while \method{} moves repeated scene elements
  coherently through \inst{} transforms.}
  \label{fig:animation_demo}
  \vspace{-2mm}
\end{figure}

\textbf{Animation.} Animation benefits from the same factorization. A standard 3DGS scene
must displace independent Gaussians; therefore, scene elements represented by many
unrelated \splats{} cannot move coherently. \figref{animation_demo} visualizes
flow maps after applying the same wind animation parameters to 3DGS and to our
instance handles. In 3DGS, splats belonging to a single flower receive
inconsistent local motion, breaking the shape of the element. With \method{}, motion is applied through \inst{} transforms, so all \splats{} in
a reused \cand{} share a coherent displacement and flowers retain their shape.
Animation implementation details are given in \secref{appendix-edit-animation}, and we refer to the supplementary video for animations.

%% file: tables/table_6_matched_budget.tex
\begin{tabular}{lcccc}
  \toprule
  Method & PSNR$\uparrow$ & SSIM$\uparrow$ & LPIPS$\downarrow$ & MB$\downarrow$ \\
  \midrule
  ContextGS   & 20.79 & 0.559 & 0.511 & 1.760 \\
  HAC++       & \best{26.51} & 0.756 & 0.299 & 1.699 \\
  OMG         & 26.25 & \best{0.770} & \best{0.271} & 1.353 \\
  \ourssmall  & 22.01 & 0.605 & 0.431 & \best{1.204} \\
  \bottomrule
\end{tabular}

%% file: tables/table_2_component_ablation.tex
\begin{tabular}{l cccc cccc}
  \toprule
  & \multicolumn{4}{c}{\ourssmall\ ($n{=}3$)}
  & \multicolumn{4}{c}{\ourslarge\ ($n{=}3$)} \\
  \cmidrule(lr){2-5} \cmidrule(lr){6-9}
  Ablation
    & PSNR$\uparrow$ & SSIM$\uparrow$ & LPIPS$\downarrow$ & MB$\downarrow$
    & PSNR$\uparrow$ & SSIM$\uparrow$ & LPIPS$\downarrow$ & MB$\downarrow$ \\
  \midrule
  Splat DC only
    & \best{$21.98_{\pm 0.07}$} & \best{$0.56_{\pm 0.00}$} & $0.45_{\pm 0.01}$ & 1.48
    & \best{$23.23_{\pm 0.02}$} & \best{$0.65_{\pm 0.00}$} & $0.34_{\pm 0.00}$ & 3.81 \\
  $+$ instance DC
    & $21.87_{\pm 0.15}$ & $0.56_{\pm 0.01}$ & $0.46_{\pm 0.02}$ & 1.45
    & \second{$23.20_{\pm 0.05}$} & \second{$0.65_{\pm 0.00}$} & $0.34_{\pm 0.00}$ & \second{3.76} \\
  $+$ LPIPS
    & \second{$21.90_{\pm 0.08}$} & \second{$0.56_{\pm 0.00}$} & \second{$0.45_{\pm 0.01}$} & \second{1.45}
    & $23.09_{\pm 0.07}$ & $0.65_{\pm 0.00}$ & \second{$0.33_{\pm 0.00}$} & 3.76 \\
  Full ($+$ adv)
    & $21.74_{\pm 0.12}$ & $0.55_{\pm 0.01}$ & \best{$0.37_{\pm 0.02}$} & \best{1.44}
    & $22.88_{\pm 0.02}$ & $0.63_{\pm 0.00}$ & \best{$0.27_{\pm 0.00}$} & \best{3.72} \\
  \bottomrule
\end{tabular}

%% file: sections/05_conclusion.tex
\section{Conclusion}
\label{sec:conclusion}

\method{} reconstructs a scene as a compact vocabulary of reusable
\cands{} together with lightweight world-space \insts{} that place
transformed copies throughout the scene. The central challenge is fitting
this shared, partly discrete structure directly from multi-view images:
each \cand{} receives supervision through many placed copies, while each
\inst{} must select and position a template. We make this optimization
practical with vocabulary-scoped \splat{} birth and relocation, world-space
\inst{} pruning, cloning, and subdivision, detached \cand{}/\inst{}
optimization, and adversarial patch supervision that preserves detail under
sharing. The resulting representation is compact, structured, editable,
and animatable. Our experiments show that this structure can reduce storage
by orders of magnitude and expose manipulation handles that flat Gaussian
fields lack, while making the expected tradeoff in image-space metrics.

\paragraph{Limitations.}
\method{} is most appropriate when the captured scene contains real
repetition. In scenes dominated by unique geometry or strong per-instance
variation, sharing becomes a poor prior: the method either underfits or
spends capacity on effectively one-off templates. The current hard
\cand{} assignment $\assign_i \in \{1, \dots, \ncands\}$ is fixed during
reconstruction, which is stable but prevents gradients from moving an
\inst{} between plausible \cands{}. Training is also more expensive than
vanilla 3DGS and many other methods because of the two-pass optimization, 
discriminator and LPIPS loss, as well as the additional density-control.

\paragraph{Outlook.}
Direct extensions include soft or learned assignments, multi-class
vocabularies for heterogeneous scenes, and generative models over instance
pose, scale, opacity, and assignment. More broadly, compositional
vocabularies are a natural target for neural reconstruction: structured
enough to edit and animate, compact enough to ship, and expressive enough
to fit real captures. \method{} is a step toward making recovered repeated
parts a first-class output, rather than an afterthought imposed on an
unstructured Gaussian field.